\documentclass[letterpaper, 10 pt, conference]{ieeeconf}  %

\usepackage{graphicx} %
\usepackage{xcolor} %
\usepackage{multirow}
\usepackage{tabularray}
\usepackage{cite}

\IEEEoverridecommandlockouts
\begin{document}

\title{Low-cost Multi-agent Fleet for Acoustic Cooperative Localization Research}
\author{Nelson Durrant, Braden Meyers, Matthew McMurray, Clayton Smith, Brighton Anderson, \\ Tristan Hodgins, Kalliyan Velasco, and Joshua G. Mangelson
  \thanks{ This work was funded by the Naval Sea Systems Command (NAVSEA), Naval Surface Warfare Center - Panama City Division (NSWC-PCD) under the Naval Engineering Education Consortium (NEEC) Grant Program under award number N00174-23-1-0005. We additionally thank NSWC-PCD for providing the initial DREW-UV CAD model, and Maria Florian from Florida State University  for the initial design of the servo housing.} 
  \thanks{N.~Durrant, B.~Meyers, M.~McMurray, C.~Smith, B. Anderson, T. Hodgins, K. Velasco, and J.G. Mangelson are at Brigham Young University. They can be reached at \texttt{\{snelsond, bjm255, mcmurram, cas314, bja1701, tah88, kalliyan, mangelson\}@byu.edu}.} 
}

\maketitle

\begin{abstract}
    Real-world underwater testing for multi-agent autonomy presents substantial financial and engineering challenges. In this work, we introduce the Configurable Underwater Group of Autonomous Robots (CoUGARs) as a low-cost, configurable autonomous-underwater-vehicle (AUV) platform for multi-agent autonomy research. The base design costs less than \$3,000 USD (as of May 2025) and is based on commercially-available and 3D-printed parts, enabling quick customization for various sensor payloads and configurations. Our current expanded model is equipped with a doppler velocity log (DVL) and ultra-short-baseline (USBL) acoustic array/transducer to support research on acoustic-based cooperative localization. State estimation, navigation, and acoustic communications software has been developed and deployed using a containerized software stack and is tightly integrated with the HoloOcean simulator. The system was tested both in simulation and via in-situ field trials in Utah lakes and reservoirs.
\end{abstract}

\section{Introduction}
\label{sec:intro}

Effective state estimation for underwater robotics is a challenging problem that is actively being addressed in academic circles. In marine environments, global positioning systems, such as GPS, are inaccessible to underwater robots and cannot be consistently relied upon for an accurate global position estimate. The majority of existing autonomous underwater vehicles (AUVs) attempt to address this problem by leveraging a combination of acoustic velocity measurements from a doppler velocity log (DVL) and inertial or heading measurements from an inertial measurement unit (IMU) or compass. While fusion of these sensors provides a strong dead reckoning solution for vehicle motion, without a global reference the localization solution will drift over time. 

One method for addressing this problem is to leverage environment features such as recognizable landmarks or terrain to bound localization drift. Much work has been done in this respect to create landmark-based simultaneous localization and mapping (SLAM) solutions, but these solutions rely on the observation of recognizable features in the environment. This is often a challenge in feature-poor underwater domains. To improve SLAM approaches in these environments, we are currently exploring methods involving multiple cooperative agents that work in tandem and share relevant information. Over the last several years, we have developed promising solutions to this problem in simulation \cite{kalliyan}; however, real-world testing of these methods at scale is a non-trivial task. One major hurdle is the high cost and proprietary hardware of commercially-available AUV systems, often making it cost prohibitive to test multi-agent operations at scale.

\begin{figure}[t]
    \centering
    \includegraphics[width=\columnwidth]{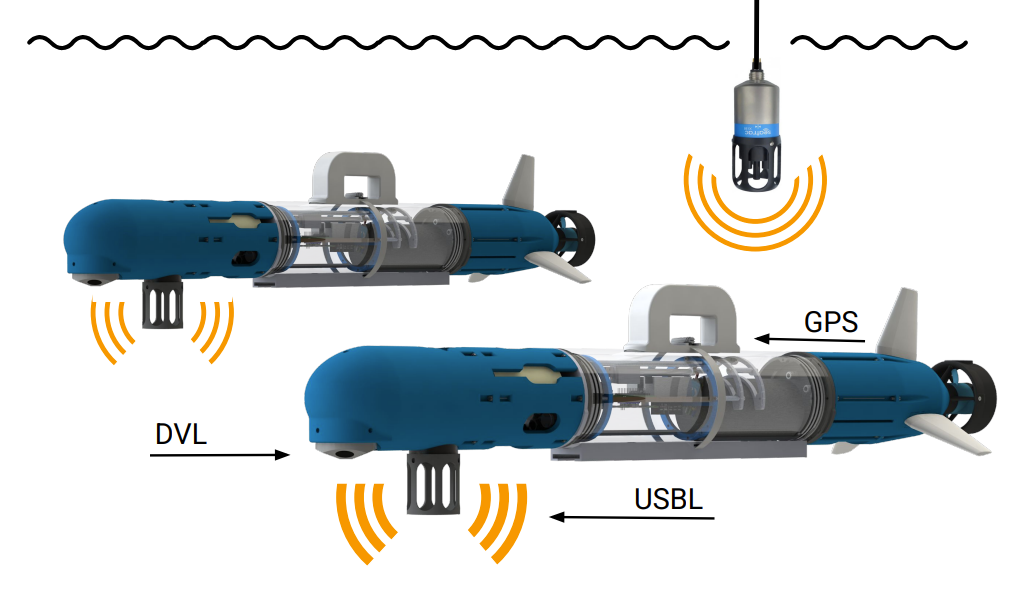}
    \caption{Configurable Underwater Group of Autonomous Robots}
    \label{fig:highlevel}
\end{figure}

To effectively address this challenge, we introduce the Configurable Underwater Group of Autonomous Robots (CoUGARs) as a low-cost, configurable AUV platform for multi-agent autonomy research. The CoUGARs fleet is composed of several AUVs referred to individually as CougUVs. In this paper, we provide an overview of our initial hardware and software design alongside results from preliminary field tests and experiments.

Our specific contributions include:
\begin{itemize}
    \item Development of a low-cost, 3D-printed, easily-configurable AUV design;
    \item Development of containerized ROS 2 localization, navigation, and acoustic communications software enabling control and operation of the CoUGARs fleet; and
    \item Presentation of field-test results demonstrating the ability of the proposed system to execute in-water field trial experiments.
\end{itemize}

The remainder of this paper is organized as follows. Section II contains a discussion of related work. Sections III and IV provide overviews of an individual CougUV's hardware and software architecture. Section V describes our initial multi-agent coordination approach using acoustics. Section VI describes both the results of testing in simulation using the HoloOcean simulator and in-situ at various lakes and reservoirs in Utah. Finally, Section VII provides a conclusion and brief discussion of future work.

\begin{figure*}[t]
    \centering
    \includegraphics[width=\linewidth]{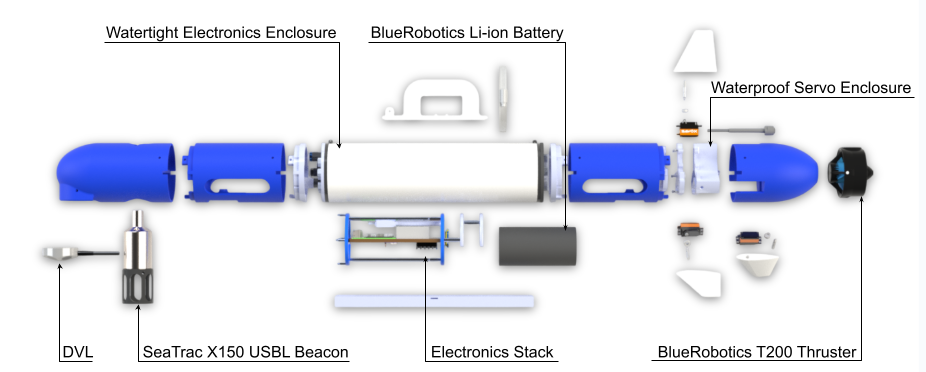}
    \caption{Exploded view of the CougUV} 
    \label{fig:cad}
\end{figure*}

\section{Related Work}
\label{sec:related_work}

Many AUV solutions are available on the market today, ranging from one-to-two man operable systems such as the Bluefin Sandshark \cite{Sandshark}, L3Harris Iver3 \cite{iver3}, or REMUS 100 \cite{Remus-100} up to large-scale systems comparable to the BlueFin 21 or REMUS 600 \cite{Remus-600}. Although these are great industrial options, their high cost, size, and proprietary hardware/software often inhibit scaling of multi-agent academic research experiments to more than a few vehicles. 

These challenges have led to the development of multiple low-cost research platforms with differing approaches and intentions. In particular, several AUV designs have been released as platforms for testing novel maneuverability capabilities. The HippoCampusX \cite{hippo} and Lancelet \cite{lancelet} designs are a good example of this, embracing a drone-like thruster configuration or a small form factor to increase agility in tight spaces. Other platforms have taken inspiration from marine animals -- the Morpheus \cite{morpheus}, SoFi \cite{sofi}, and U-CAT \cite{ucat} designs explore alternative AUV actuation systems mimicking a tuna, fish, and turtle respectively. The LoCO \cite{loco} and MeCO \cite{meco} designs, equipped with cameras and user communication devices, serve as interesting low-cost and medium-cost AUV options for experimenting with underwater human-robot interaction (UHRI).

All of these AUV design options carve out a specific research niche or implement experimental design elements, but the increased access to rapid prototyping with the wide-spread adoption of 3D-printing and open-source robotics software frameworks such as MOOS and ROS have led to several designs that that take a more general-purpose approach. Some notable platforms that fall into this category are the BlueJacket UUV \cite{bluejacket}, the ALPHA AUV \cite{alpha}, and the Spurdog AUV \cite{turrisi2024spurdog}. All of the above solutions should be acknowledged for their 3D-printed, commercial-off-the-shelf (COTS) approach to building a general-purpose, low-cost AUV platform. The BlueJacket UUV presents a drive-shaft servo waterproofing solution for a 3D-printed flooded aft section \cite{bluejacket}, while the Spurdog opts for the Trident aft section \cite{turrisi2024trident}, a hybrid design with both flooded and sealed components. On the software side, the ALPHA and Spurdog AUVs both include a full navigation and simulation stack. The ALPHA AUV is based in ROS 1 and integrated with the Stonefish simulator, while the Spurdog AUV leverages MOOS-IvP for both navigation and simulation \cite{alpha} \cite{turrisi2024spurdog}.

In this work, we present our development of the CoUGARs platform, specifically designed to support multi-agent acoustic localization and autonomy experiments. Our proposed CougUV (based off the DREW-UV system developed at NSWC Panama City \cite{dubyoski2022DrewUV}) adds another low-cost, easily-constructed, and configurable AUV to the larger academic community.

\section{Hardware Design}
\label{sec:hardware_design}

The CougUV is designed for ease of assembly and modification, and features a Raspberry Pi 5 and Teensy 4.1 microcontroller along with a COTS DVL, ultra-short-baseline (USBL) acoustic array/transducer, GPS, IMU, and depth sensors. An exploded view of the CAD design is provided in Fig.~\ref{fig:cad}. As of May 2025, the base design costs less than \$3,000 USD, and two CougUVs have been constructed and tested by the BYU FRoSt Lab (with plans to construct three more). We provide additional detail on the electrical and hardware design in the following sub sections.

\subsection{Electrical System}

\begin{figure}[t]
    \centering
    \includegraphics[width=\columnwidth]{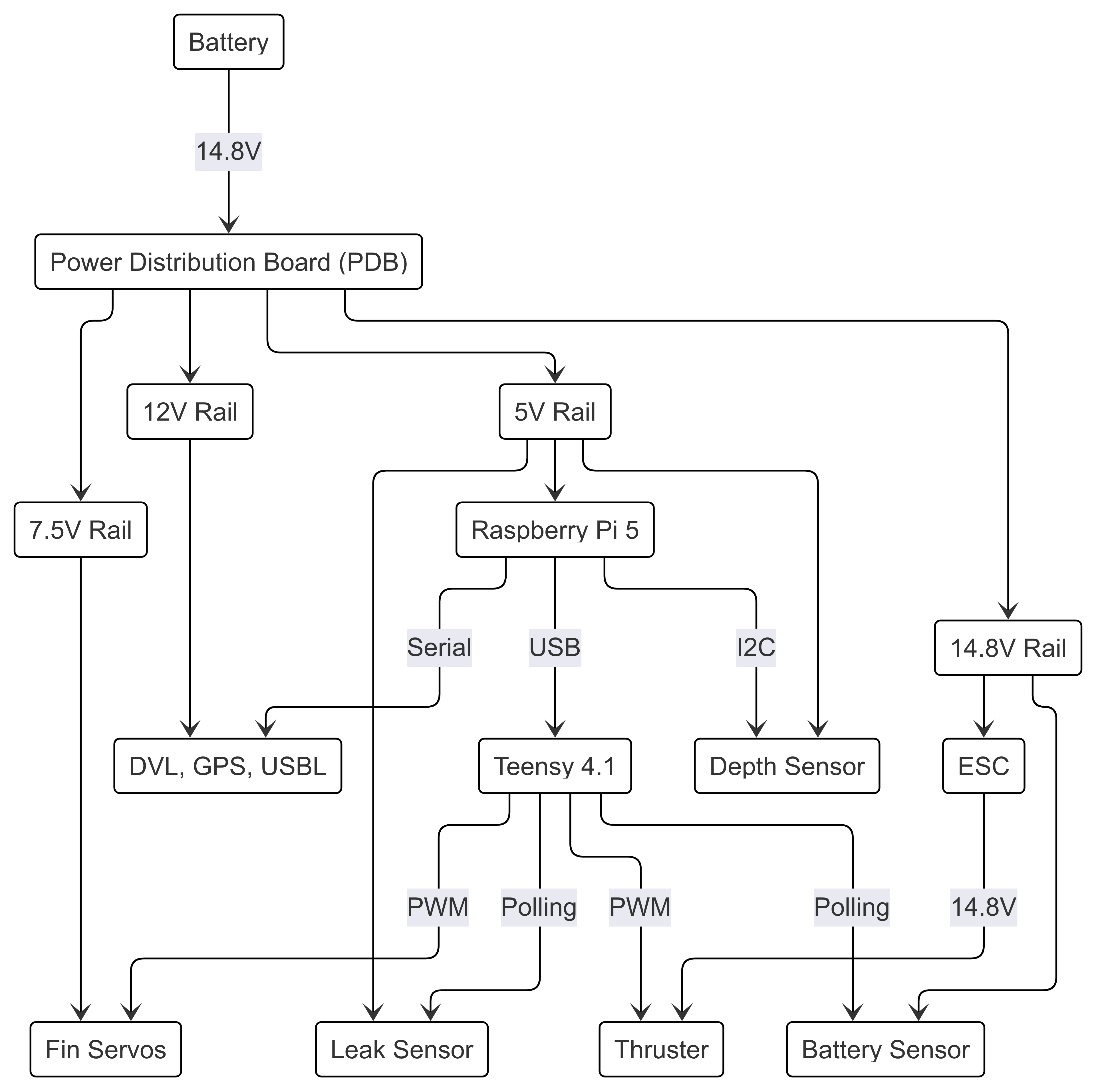}
    \caption{Electrical System Diagram}
    \label{fig:Electronics Diagram}
\end{figure}

An overview of the primary electrical system for the CougUV is shown in Fig.~\ref{fig:Electronics Diagram}.
The CougUV's electrical system leverages a Raspberry Pi 5 for high-level processing and a Teensy 4.1 microcontroller for real-time sensor interfacing and actuator control. The Raspberry Pi hosts and runs a dedicated Docker container pre-installed with the dependencies needed to run our ROS 2 packages and software, including the state estimator and mission planner, while the Teensy is responsible for generating the pulse-width modulation (PWM) signals to control the thruster and fin servos and polling simple sensors such as the leak sensor and battery monitor.

Power is supplied by a 14.8V, 18Ah lithium ion battery pack, which is regulated by a custom power distribution board (PDB). The PDB provides stable 12V, 7.5V, 5V, and 3.3V rails for the various sensors and motors onboard.  Communication with the main sensors is handled via serial (DVL, GPS, USBL), USB (IMU), and I2C (depth sensor) interfaces, which are all routed to the Raspberry Pi. This centralized architecture simplifies the software interfaces and allows the Teensy to be dedicated to time-sensitive control of the fins and thruster.

\subsection{3D Printing and Assembly}

The majority of the non-COTS parts of the CougUV design are intended and designed to be manufactured using fused deposition modeling (FDM) 3D-printers and Polyethylene Terephthalate Glycol (PETG) filament. Combined with the wide-spread availability and adoption of 3D-printing, this design approach allows for both quick vehicle customization for different sensor payloads and rapid prototyping of those same designs. All FDM 3D-printed parts are printed without infill for waterproofing and buoyancy considerations and are easily attached or replaced using metric screws.

\subsection{Buoyancy and Ballast}
The CougUV is designed to be slightly positively buoyant to minimize power consumption and allow the CougUV to float to the surface in the case of loss of control or power. Care was taken to optimize the centers of mass and buoyancy to generate a restoring force that stabilizes roll.

To save space and weight, multiple segments of the vehicles are hollow when printed and then filled with rigid polyurethane foam. The foam is mixed and then poured to fill the hollow section of the 3D print in order to create custom foam shapes. We selected a 12 lbs/ft$^3$ density polyurethane foam that is rated for depths up to 91 meters.
Ballast weight is added to a weight tray located below the vehicle, as seen in Fig.~\ref{fig:cad}. The weight tray is designed such that weight can be easily added, removed, or adjusted in the field depending on the density and salinity of the water. 

\subsection{Waterproofing}

\begin{figure}[t]
    \centering
    \includegraphics[width=\columnwidth]{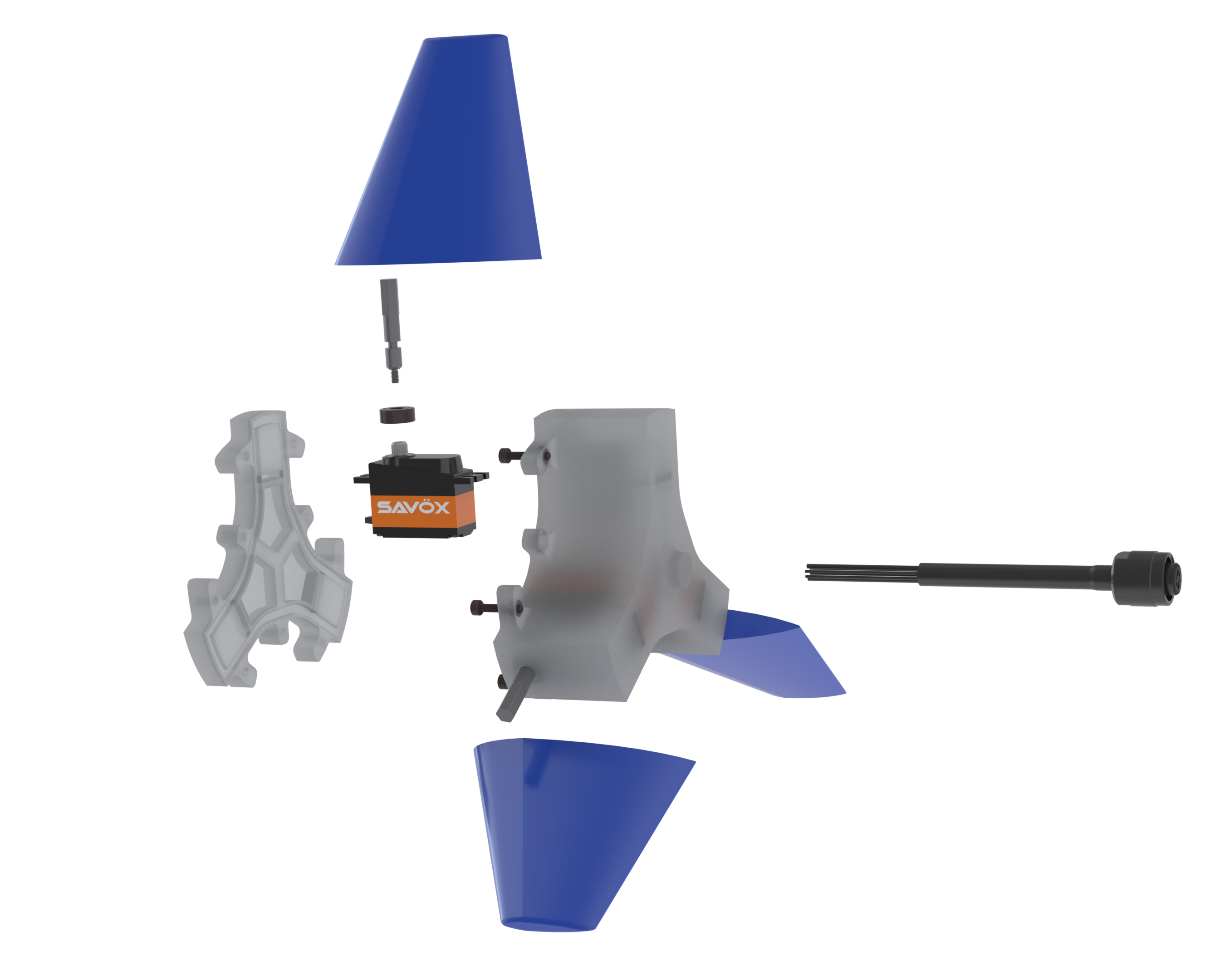}
    \caption{Exploded view of the CougUV's watertight servo pressure housing }
    \label{fig:fins}
\end{figure}

The original DREW-UV design was rated to a depth of a few meters, with the limiting factor being the servo motors that actuate the fins. The CougUV has an upgraded design with a stereolithography (SLA) 3D-printed waterproof pressure housing designed to protect low-cost servos at depth. We have successfully pressure tested the servo housing and demonstrated its ability to withstand pressure equivalent to more than 60 meters in depth.

The housing holds three Savox servos. The enclosure for the servo motors features a wet-connect Blue Trail Engineering cable that provides power and PWM signals to the servos. The lower end of a stainless steel hex standoff is machined to a cylindrical shaft with a groove to hold a small o-ring as seen in Fig.~\ref{fig:fins}. The shaft is supported by a ball bearing inside the housing and the o-ring contacts the smoothed surface of the SLA housing to create the seal. The fins attach on the hex end of the shaft and are locked from rotating by the hex pattern. This housing was designed to minimize manufacturing time and can be replaced simply in the event of failure. 

The electrical system and battery are housed within a standard Blue Robotics four-inch acrylic tube enclosure. With the improved servo housing, this enclosure is the limiting factor for depth, placing the theoretical depth rating of the CougUV at 60 meters, although it has not yet been tested to that depth. 

\section{Software Architecture}
\label{sec:software_architecture}

\begin{figure}[t]
    \centering
    \includegraphics[width=\columnwidth]{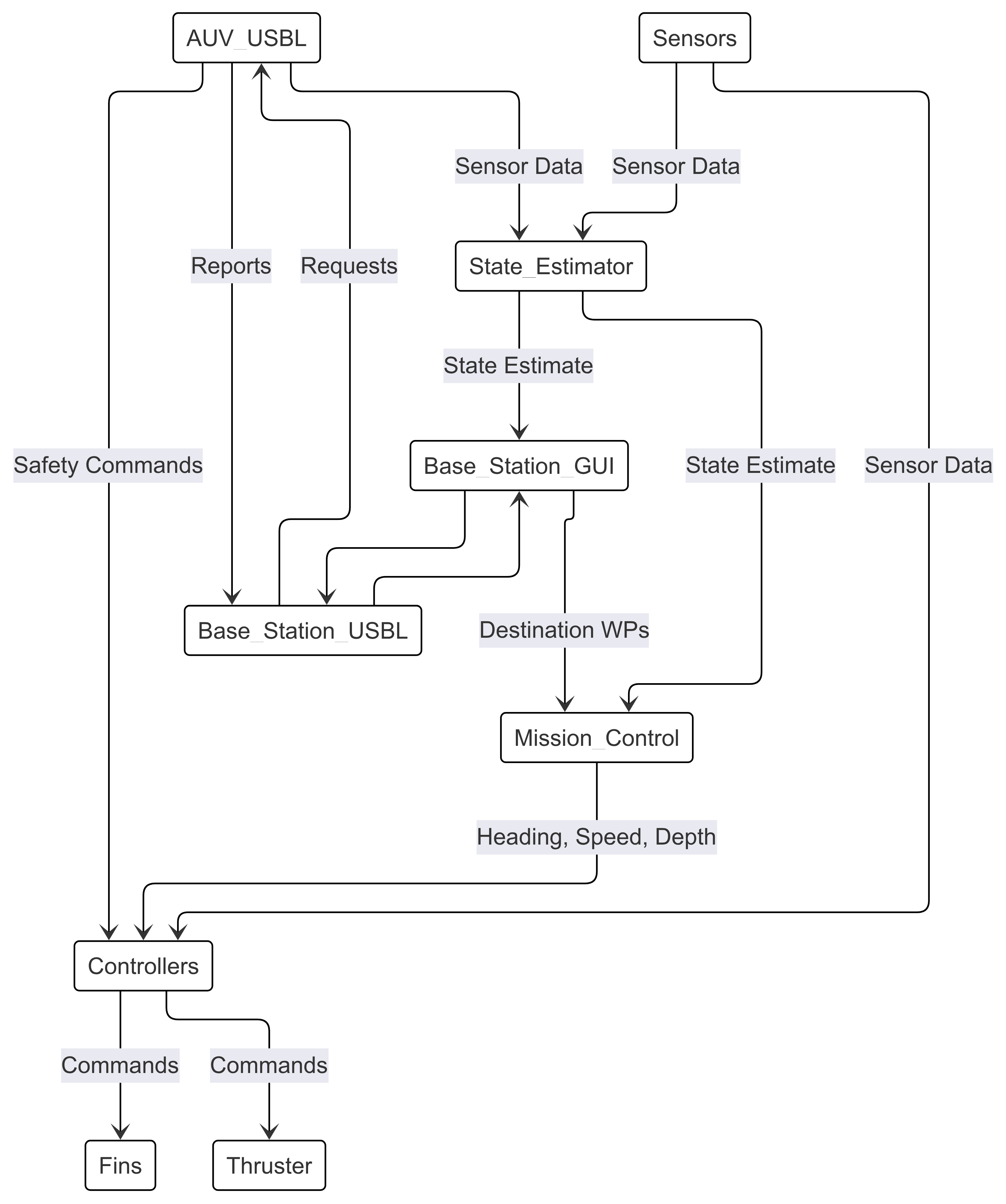}%
    \caption{Software Architecture Diagram}%
    \label{fig:software}
\end{figure}

The CougUV software stack is implemented using ROS 2 \cite{ros2} and containerized using Docker \cite{docker} for ease of development, simulation, and deployment. An overview of the CougUV software architecture is shown in Fig.~\ref{fig:software}. The software stack contains a high-level mission planner, factor-graph-based state estimator, controllers, and hardware drivers. We provide additional detail on these items in the following sub sections.

\subsection{State Estimation}

\begin{figure}[t]
    \centering
    \includegraphics[width=\columnwidth]{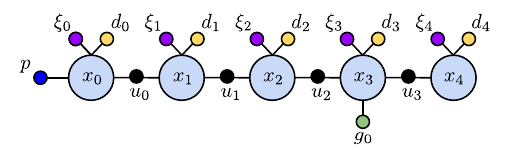}
    \caption{A single-agent pose graph, where $x_i$ indicates the pose at each timestep $i$, and the factors are indicated by $p_0$ (prior), $u_i$ (DVL), $\xi_i$ (IMU), $d_i$ (depth), and $g_i$ (GPS). }
    \label{fig:single_agent_fg}
\end{figure}

The CougUV software stack performs state estimation for AUVs using a GTSAM-based factor graph \cite{gtsam}. The single-agent pose graph is shown in Fig.~\ref{fig:single_agent_fg}. In this representation, variable nodes correspond to six degree-of-freedom (DOF) vehicle poses at discrete time steps, while factor nodes encode probabilistic constraints derived from sensor measurements. We incorporate unary GPS factors (when at the surface) to constrain the translation of robot pose, DVL factors to constrain the 6-DOF relative motion between poses, IMU factors for orientation, and depth factors for the $z$ component of the robot pose. The pose graph is initialized at the agent’s local origin and constrained using a unary prior factor. The graph is evaluated using the iSAM2 optimizer \cite{isam2}. 

We implement time synchronization and message queuing to associate sensor measurements with the correct pose nodes in the factor graph. Each pose node is timestamped using its corresponding DVL measurement. As sensor messages arrive, they are queued along with their timestamps. For each new pose node, we compare the time offsets between queued sensor messages and both the current and previous pose node timestamps to determine the best match. Measurements are then assigned to the closest pose or discarded if no suitable match is found. This nearest-neighbor strategy ensures accurate temporal alignment despite varying sensor rates and delays.

The factor graph has been validated both in real-time deployments and through post-processed mission data.

\subsection{Control Systems}

\begin{figure*}[t]
    \centering
    \includegraphics[width=2\columnwidth]{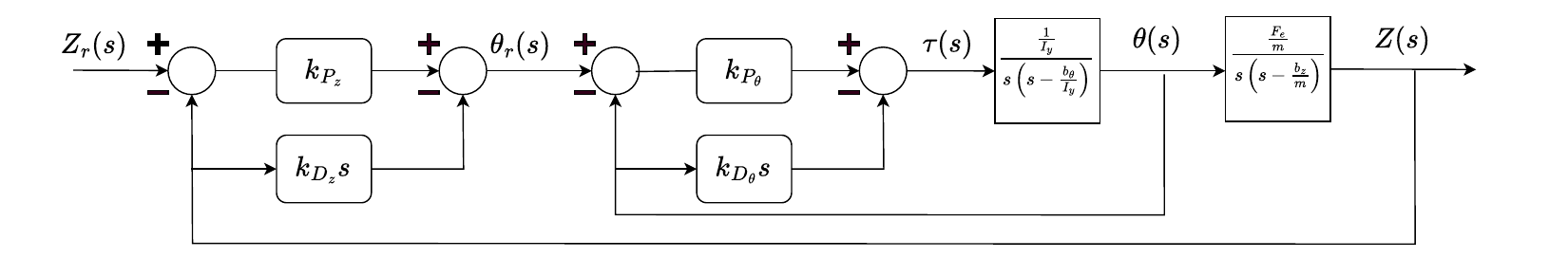}
    \caption{Nested control flow for Pitch/Depth}
    \label{fig:Depth Control Diagram}
\end{figure*}

Real-time control of the CougUV is divided into three separate subsystems: speed, heading, and pitch/depth. The vehicle is actuated via three fins and a single thruster. For simplicity, we treat these subsystems as being decoupled and non-interacting. While in reality this is not the case (since for example, you cannot control heading or pitch using the fins when the vehicle is not moving forward), this simplifies the control system implementation and works well in practice.

We actuate the thruster to control speed, the top fin to control heading, and the port and starboard fins to control pitch and depth. We linearize our heading and pitch/depth controls about a constant forward speed. 

State estimation inputs are derived from depth and Attitude Heading and Reference (AHRS) sensor data. These measurements are globally fixed and calibrated and then passed directly to the controller. Commands from the high-level mission manager are provided as inputs to the control system. While heading and speed follow the standard PID pattern, we nest pitch and depth as depth cannot be directly controlled. The pitch/depth PD controller diagram shown in Fig.~\ref{fig:Depth Control Diagram} was based on work from Thor Fossen \cite{fossen}. 

The outer depth control loop takes the commanded depth $Z_r(s)$ and compares it with measured depth $Z(s)$, with the error processed by a PD controller ($k_{P_z}$, $k_{D_z}s$) to output the reference pitch $\theta_r(s)$. This $\theta_r(s)$ then feeds the inner pitch control loop. Here, the reference pitch is compared to measured pitch $\theta(s)$, and the error is processed by another PD controller ($k_{P_\theta}$, $k_{D_\theta}s$) to generate the control torque $\tau(s)$. The system's pitch dynamics, represented by an appropriate transfer function, translate $\tau(s)$ to $\theta(s)$. In these dynamics, $I_y$ is the moment of inertia about the pitching axis, and $b_\theta$ is the pitch damping coefficient. Finally, the depth dynamics, also represented by an appropriate transfer function, relate $\theta(s)$ to the resulting depth $Z(s)$. For the depth dynamics, $m$ represents the vehicle's mass, $b_z$ is the depth damping coefficient, and $F_e$ is the constant force from the thruster around which the system dynamics are linearized. 
This nested architecture enables precise depth control by manipulating pitch.

\subsection{Mission Planning and Operations}
The low-level AUV heading, speed, and depth controllers are intentionally decoupled from the high-level mission planner to enforce system modularity. The controllers simply subscribe to desired heading, speed, and depth commands on a specified ROS 2 topic, allowing for straightforward integration with any high-level planner capable of publishing them. Our current implementation makes use of the MOOS-IvP Helm \cite{Benjamin2010NestedAF} mission controller, which interfaces with the system via a custom software bridge between the ROS 2 and MOOS middlewares. Waypoints are passed as behavior files into the MOOS-IvP mission planner, which then determines and publishes the desired heading, speed, and depth of the AUV given its current state estimate and destination. For future iterations of the vehicle, a custom waypoint-based mission planner is also under development.

\section{Multi-Agent Coordination and Communications}
\label{sec:multiagent_comms}

Our main goal in developing the CoUGARs fleet is to enable large-scale multi-agent autonomy experiments and field trials. We have begun to leverage these capabilities in the field and on real-world hardware. Our current experiments are centralized in nature and consist of a BlueROV as the lead agent and two CougUVs as followers. The lead agent serves as the message instigator during multi-agent missions. Each CougUV, in its current configuration, is equipped with a BluePrint Seatrac X150 acoustic USBL beacon \cite{seatrac} for sub-sea communications.  We also deploy a static USBL beacon for mission monitoring and control.

Because seawater attenuates high-frequency electromagnetic waves, traditional communication methods like WiFi and radio don't propagate underwater beyond a few centimeters. As a result, AUVs primarily use acoustic communications (comms) for sub-surface data exchanges. Nonetheless, acoustic comms come with significant limitations: low data rates and high latency (30 bytes every 4 seconds), unreliable transmissions, multi-path interference, and limited bandwidth (which causes signal collisions).

To avoid message collisions, acoustic comms during a mission are instigated through the lead agent. The lead agent requests a status update from each vehicle one at a time in a loop. If a response is not received within four seconds of requesting, the agent continues to ping the next follower in the rotation. Each update includes the the full 6-DOF position estimate, depth, and a status bitmask, amounting to a total of 28 bytes (within the 30 byte packet size limit). The packet size may be further reduced with a message compression scheme, which we leave to future development of the CoUGARs system.

We keep a human operator in the loop during operation of multi-agent missions through a static acoustic beacon. The human operator can monitor the status of each vehicle through a custom graphical user interface. Currently, the operator may issue acoustic commands to abort a mission, but we plan to expand the acoustic tasking to include a variety of behaviors including way-point following, surfacing, and loitering in the near future.

In addition to acoustic comms, the X150 beacon provides each agent with a USBL transducer array from which the angle of arrival of an incoming signal may be computed. This choice was made to support research in passive cooperative localization schemes such as \cite{kalliyan}.

\section{Testing and Results}
\label{sec:results}

\subsection{Simulation}
New software was tested using the HoloOcean simulator \cite{10638434} before deployment in the field. We have developed a ROS 2 bridge for HoloOcean that enables us to conduct both software and hardware-in-the-loop testing of the CougUV system. In these tests, noisy sensor data from the simulation is routed using ROS 2 to the CougUV's computing system, and the physical actuator control commands are then sent back to the simulator to enable verification of planning, control, and perception algorithms before testing in the field. Further details will be outlined in a separate publication. 

\subsection{Field Testing}

\begin{figure}[t]
    \centering
    \includegraphics[width=\columnwidth]{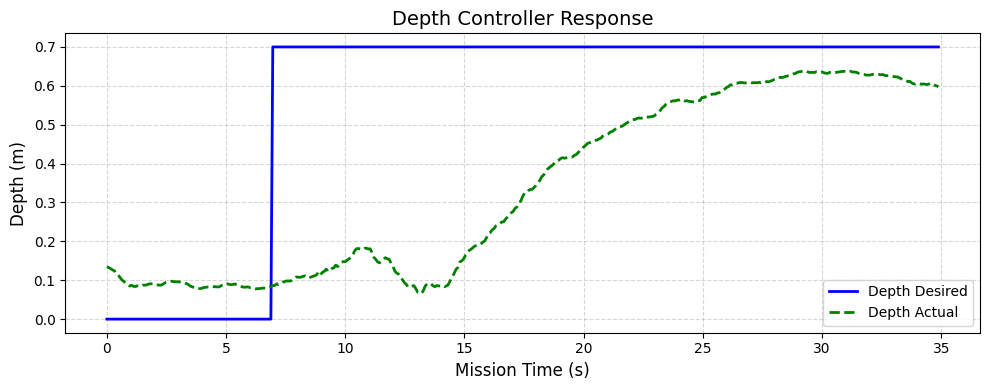}
    \includegraphics[width=\columnwidth]{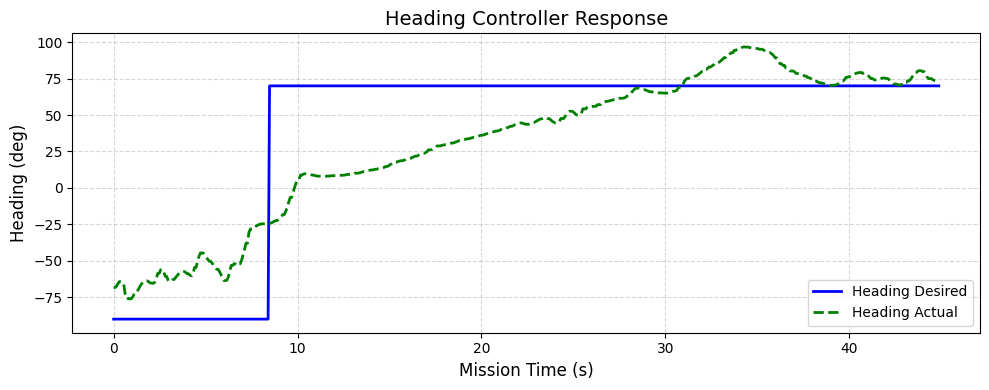}
    \caption{CougUV depth and heading controller responses during a mission in Bear Lake, Bear Lake, UT, Jul. 2025}
    \label{fig:bear-lake-control}
\end{figure}

\begin{figure}[t]
    \centering
    \includegraphics[width=\columnwidth]{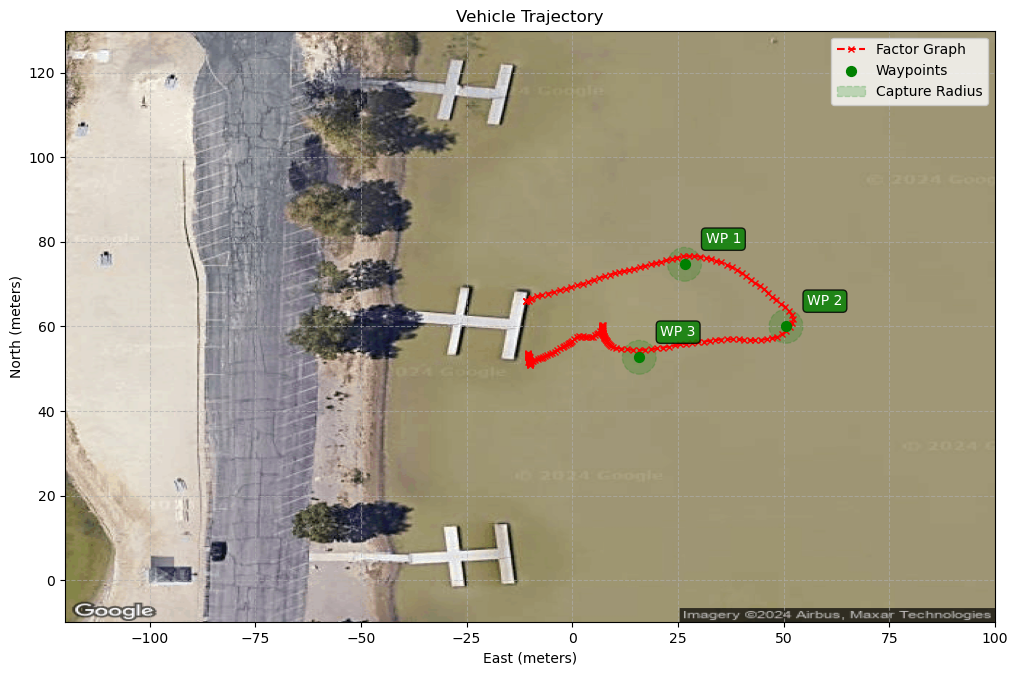} 
    \caption{A CougUV's reported trajectory from following a successful waypoint mission in Utah Lake, Provo, UT, Nov. 2025}
    \label{fig:utah-lake-trajectory}
\end{figure}

\begin{figure}[t]
    \centering
    \includegraphics[width=\columnwidth]{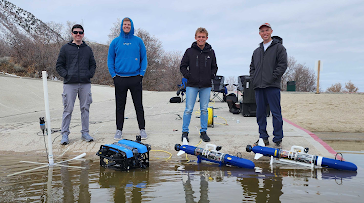}
    \caption{Two CougUVs, modified BlueROV, and base-station USBL (mounted on PVC stand) during a multi-agent field test at Spanish Oaks Reservoir, Spanish Fork, UT, Apr. 2025}
    \label{fig:test}
\end{figure}

Along with simulation in HoloOcean, the CougUV's hardware and software systems have been validated in over twenty field operations in Utah lakes and reservoirs, along with additional testing in our lab's test tank and the dive pool at Brigham Young University. The primary objectives of these trials were to: (1) tune and verify the low-level controllers, (2) test and evaluate the navigation and state estimation software, and (3) demonstrate multi-agent acoustic communication and data collection capabilities. 

A fundamental requirement for AUV autonomy is robust low-level control of the vehicle's position and velocity. Heading and depth controllers were first tuned in simulation and at the dive pool before deployment in the field. In Fig.~\ref{fig:bear-lake-control}, we show the vehicle's response to two separate step inputs during a mission in Bear Lake, UT -- a dive from the surface to a depth of 0.7 meters and a heading change from 90 to -90 degrees. The vehicle is shown to have struggled initially on the dive as the thruster tipped intermittently up and out of the water with the commanded forward pitch, but quickly approaches the desired depth once submerged.

We also validated the performance of the factor-graph-based state estimator and navigation logic. In Fig.~\ref{fig:utah-lake-trajectory}, we show the trajectory of a CougUV (as calculated by the state estimator fusing DVL, IMU, depth, and GPS measurements) during the completion of a multi-waypoint mission in Utah Lake, UT. In the test, the vehicle was given three successive GPS points and successfully navigated to each of them within the specified goal tolerance radius included in the mission plan.

Additionally, we have demonstrated the capability to repeatedly and accurately collect synchronized data from the USBL and DVL, our main sensor payloads, and record inter-vehicle acoustic communications using the base station USBL. Fig.~\ref{fig:test} shows a picture of the setup used to support our multi-agent field testing operations.

\section{Conclusion}
\label{sec:conclusion}

Testing multi-agent autonomy and coordination solutions at scale is limited by the cost, complexity, and availability of suitable AUV systems. As a response to this challenge, the CoUGARs platform provides a low-cost, configurable AUV built from easily-acquirable COTS and 3D-printed parts. This design allows for the rapid and straightforward integration of specialized sensor payloads; our current configuration includes a DVL and USBL to specifically support multi-agent acoustic localization research. We have successfully developed containerized state estimation, control, and acoustic communications software and integrated it tightly with the HoloOcean simulator for rapid testing and iteration. The entire system has been verified in over twenty field operations conducted in Utah lakes and reservoirs.

We acknowledge there is still much work to be done in creating a truly robust multi-agent acoustics platform. The simple request-response protocol and low data exchange rate of our current setup could be significantly improved by implementing a message compression scheme. Testing has also been limited to relatively calm, freshwater environments with no more than three agents. Future developments will include the addition of features such as a mast for better GNSS, WiFi, and radio reception at the surface, improvements to the acoustic command protocol to allow for more effective management of multiple agents, and the expansion of our current CoUGARs fleet from two to five vehicles.

We plan to release hardware and software documentation for the CoUGARs and CougUV systems at the following link: https://frost-lab.gitbook.io/cougars/.

\bibliographystyle{IEEEtran}
\bibliography{ref}
\end{document}